\documentclass{article}
\usepackage{myCommands}
\usepackage{times}
\usepackage{graphicx} 
\usepackage{subfigure} 
\usepackage{natbib}
\usepackage{algorithm}
\usepackage{algorithmic}
\usepackage{hyperref}

\usepackage[accepted]{icml2014}

\icmltitlerunning{An Analysis of State-Relevance Weights and Sampling
Distributions on RALP Approximation Accuracy}

\begin{document} 

\twocolumn[
\icmltitle{An Analysis of State-Relevance Weights and Sampling Distributions
on $L_1$-Regularized Approximate Linear Programming Approximation Accuracy}

\icmlauthor{Gavin Taylor}{taylor@usna.edu}
\icmlauthor{Connor Geer}{}
\icmlauthor{David Piekut}{}
\icmladdress{United States Naval Academy,
  572M Holloway Rd., Stop 9F, Annapolis, MD  21402-5002}

\icmlkeywords{Reinforcement Learning, Feature Selection, L1 Regularization,
Value Function Approximation}

\vskip 0.3in
]

\begin{abstract}
  Recent interest in the use of $L_1$ regularization in the use of value
  function approximation includes Petrik et al.'s introduction of
  $L_1$-Regularized Approximate Linear Programming (RALP).  RALP is unique among
  $L_1$-regularized approaches in that it approximates the optimal value
  function using off-policy samples.  Additionally, it produces policies which
  outperform those of previous methods, such as LSPI.  RALP's value function
  approximation quality is affected heavily by the choice of state-relevance
  weights in the objective function of the linear program, and by the
  distribution from which samples are drawn; however, there has been no
  discussion of these considerations in the previous literature.  In this
  paper, we discuss and explain the effects of choices in the state-relevance
  weights and sampling distribution on approximation quality, using both
  theoretical and experimental illustrations.  The results provide insight not
  only onto these effects, but also provide intuition into the types of MDPs
  which are especially well suited for approximation with RALP.
\end{abstract}

\section{Introduction}
\label{sec:intro}

In recent years, the Reinforcement Learning community has paid considerable
attention to creating value function approximation approaches which perform
automated feature selection while approximating a value
function~\cite{kolter09a,johns10,mahadevan12,liu12}.  This approach frees
researchers from hand-selecting and -tuning feature sets, while greatly
increasing approximation accuracy.  One of these approaches, $L_1$-Regularized
Approximate Linear Programming (RALP)~\cite{icml10,taylor12} is unique in that
it results in an approximation of the optimal value function and makes use of
off-policy samples.

However, important aspects of RALP have not been fully explored or explained.
In particular, the objective function of the linear program offers an
opportunity to create a better approximation of the value function in some
regions of the state space at the expense of others.  Optimal policies of many
realistic reinforcement learning problems heavily traffic some parts of the
state space while avoiding others, making an understanding of this flexibility
useful.

Therefore, in Section \ref{sec:rho}, we derive a new error bound for the RALP
approximation, which is tighter than those presented by Petrik et
al.~\yrcite{icml10} and provides new insight into the result of changing the
state-relevance weights in the objective function.  In addition, this bound
provides an insight into the types of MDPs particularly well suited to the
RALP approach.  Finally, this section provides evidence that rather than
weighting all states equally, as was done by Petrik et al.~\yrcite{icml10} and
Taylor and Parr~\yrcite{taylor12}, states should instead be weighted in
proportion to the stationary distribution under the optimal policy.


Additionally, if the sampling distribution is not uniform across the state
space, the approximation can be heavily affected.  In realistic domains,
sampling is rarely uniform.  Therefore, for RALP to be appropriate for these
domains, it is important to address this lack of analysis and understand how
the approximation is likely to be altered due to the problem's available
sampling scheme.

Section \ref{sec:samp} then discusses the impact on the approximation if
learning is performed using samples drawn other than uniformly from the state
space.  We demonstrate that sampling from a distribution acts as a \emph{de
facto} alteration of the objective function.  We also discuss the effects of
sampling distributions in light of the bounds from Section \ref{sec:rho}.


The intuition provided by Sections \ref{sec:rho} and \ref{sec:samp} are then
demonstrated experimentally in Section \ref{sec:experiments}.  Using a simple,
easily visualized domain, we demonstrate the effect of the various parameters
on approximation quality.  

\section{Notation and Problem Statement}
\label{sec:notation}

In this section, we formally define Markov decision processes and linear value
function approximation.  A \emph{Markov decision process} (MDP) is a tuple
$(\sS,\sA,P,R,\gamma)$, where $\sS$ is the measurable, possibly infinite set
of states, and $\sA$ is the finite set of actions. $P: \sS \times \sS \times
\sA \mapsto [0,1]$ is the transition function, where $P(s'|s,a)$ represents
the probability of transitioning from state $s$ to state $s'$, given action
$a$. The function $R: \sS \mapsto \Re$ is the reward function, and $\gamma,$ a
number between 0 and 1, is the discount factor, representing the comparative
desire for reward at the current time step to the desire for reward at the
next time step.

We are concerned with finding a value function $V$ that maps each state
$s\in\sS$ to the expected total $\gamma$-discounted reward for the
process.  Value functions can be useful in creating or analyzing a policy
$\pi: \sS \times \sA \rightarrow [0,1]$ such that for all
$s\in\sS$, $\sum_{a\in\sA} \pi(s,a) = 1$.  The transition and reward
functions for a given policy are denoted by $P_\pi$ and $R_\pi$.  We denote
the Bellman operator for a given policy as $T_\pi,$ and the max
Bellman operator simply as $T$.  That is, for some state $s\in\sS$: 
\[
T_\pi V(s)=R(s)+\gamma\int_{\sS}P(ds'|s,\pi(s))V(s')
\]
\[
TV(s)=\max_{\pi\in\Pi}T_\pi V(s).
\]
We additionally denote the Bellman operator for selecting a particular action
$a$ as
\[
T_a V(s)=R(s)+\int_{\sS}P(ds'|s,a)V(s').
\]

The optimal value function $V^*$ satisfies $T V^*(s) = V^*(s)$ for all
$s\in\sS$.

For simplicity, in this paper we will assume no noise exists in the MDP;
results can be easily extended to noisy domains using Taylor and
Parr's~\yrcite{taylor12} approach of local smoothing.

Sets of samples, therefore, are defined as $\Sigma \subseteq \{(s,a,r,s' |
s,s'\in\sS, a\in\sA\}$, where $s'$ is the state the agent arrived at given
that it started in state $s$ and took action $a$, and $r=R(s)$.  An individual
sample in the set $\Sigma$ will be denoted $\sigma$, and an element of a
sample $\sigma$ will be denoted with superscripts; that is, the $s$ component
of a $\sigma=(s,a,r,s')$ sample will be denoted $\sigma^s$.


We focus on linear value function approximation for discounted
infinite-horizon problems, in which the value function is represented as a
linear combination of possibly nonlinear basis functions (vectors). For each
state $s$, we define a vector $\Phi(s)$ of features. The rows of the basis
matrix $\Phi$ correspond to $\Phi(s)$, and the approximation space is
generated by the columns of the matrix. That is, the basis matrix $\Phi$, and
the approximate value function $\hat V$ are represented as:
\[
\Phi = \begin{pmatrix} - & \Phi(s_1) & - \\  &
\vdots & \end{pmatrix} \qquad \hat V = \Phi w .
\]
This form of linear representation allows for the calculation of an
approximate value function in a lower-dimensional space, which provides
significant computational benefits over using a complete basis; if the number
of features is small and the environment is noisy, this framework can also
guard against overfitting any noise in the samples.

If we define $\Phi$ to be overcomplete, with potentially far more features
than sampled states, then to receive the above benefits we must perform
feature selection. In this process, a few features are chosen from the set, the
span of which will represent the available linear approximation space.  We can
use $L_1$ regularization to calculate a sparse $w$, in which nearly all
features receive a weight of 0, thereby performing automated feature selection.

\section{Previous Work}
\label{sec:previous}

RALP was introduced by Petrik et al.~\yrcite{icml10} to extend the
capabilities of the linear programming approach to value function
approximation~\cite{depenoux63,schweitzer85,deFarias03}.  Given a set of
samples $\Sigma$, the linear program is defined as follows:
\begin{equation}
\label{eqn:ralp}
\begin{array}{rl}
\displaystyle\min_{w} &\rho^T \Phi w\\
\mbox{s.t.} &T_{\sigma^a}\Phi(\sigma^s) w\leq
\Phi(\sigma^s) w ~~\forall \sigma\in\Sigma \\ &\norm{w}_{-1}
\leq \psi,
\end{array}
\end{equation}
where $\rho$ is a distribution, which we call the state-relevance weights, in
keeping with the (unregularized) Approximate Linear Programming (ALP)
terminology of de Farias and Van Roy~\yrcite{deFarias03}.
$\norm{w}_{-1}$ is the $L_1$ norm of the vector consisting of all weights
excepting the one corresponding to the constant feature.

This final constraint, which contributes $L_1$ regularization, provides
several benefits.  First, regularization in general ensures the linear program
is bounded, and produces a smoother value function.  Second, $L_1$
regularization in particular produces a sparse solution, producing automated
feature selection from an overcomplete feature set.  Finally, the sparsity
results in few of the constraints being active, speeding the search for a
solution by a linear program solver, particularly if constraint generation is
used.

Other techniques have used $L_1$ regularization in similar ways.
LARS-TD~\cite{kolter09a} and LC-MPI~\cite{johns10} both approximate the fixed
point of the $L_1$-regularized LSTD problem.  Mahadevan and
Liu~\yrcite{mahadevan12} introduced the use of mirror descent, which has a
computation complexity which allows it to be better suited than many other
approaches for online reinforcement learning problems.  These above approaches
are most reliable when samples are collected on-policy.  Liu et
al.~\yrcite{liu12} introduced RO-TD, which converges to an approximation of
the value function of a given policy, even when trained on off-policy samples.

In contrast to the above approaches, RALP provides an approximation to the
value function of the \emph{optimal} policy, even when samples are drawn from
non-optimal or random policies.  Approximations produced by RALP have bounded
error, and have performed well experimentally in comparison to other
approaches.  Finally, in noisy domains, the well-known weakness of linear
programming approaches to value function approximation can be mitigated or
eliminated using local smoothing~\cite{taylor12}.

This previous work on RALP has largely ignored the state-relevance weights
$\rho$ in the objective function, setting $\rho=\bone$ without discussion.
However, a change in the objective function would obviously affect the
solution of the linear program.  In certain practical situations this would be
useful to understand.
For example, consider the task of calculating a value function for an aircraft
in flight.  The space of possible flight attitudes, velocities, etc. is very
large.  However, the percentage of this space trafficked in non-catastrophic
flight is small; it is likely worthwhile to improve the approximation quality
in this relevant portion, at the expense of accuracy in the remainder of the
space.

Some previous results exist regarding the effects of changing the
state-relevance weights in the closely-related ALP~\cite{deFarias03}.
However, the assumptions and types of appropriate problems are very different
between ALP and RALP, making these previous results insufficient.  First, ALP
assumes a sample is drawn from every state-action pair, an assumption which is
not required for RALP.  This means it was not necessary with ALP to consider
the behavior of the approximation between samples or in a continuous space.
Furthermore, it was not necessary to consider the effects of sampling
distributions at all.  This assumption was later weakened by a followup
paper~\cite{deFarias04}, but not to a degree necessary for large or continuous
problems, particularly when large numbers of samples are not available.  The
second difference is ALP is unregularized, simplifying the definition of the
feasible space of the linear program.  Despite these differences, these
previous results will serve as a useful guide.

Additionally, previous work does not cover the effects of sampling schemes on
RALP.  If states are sampled heavily in one portion of the state space, the
linear program will choose a solution which tightens constraints in that
portion of the space over others.  In realistic settings, it can be difficult
to sample uniformly, making it especially important to understand the effect
of other sampling distributions on the resulting approximate value function.

The remainder of this document fills in these gaps in the previous work.

\section{State-Relevance Weights}
\label{sec:rho}

The theoretical results presented on RALP in the literature thus far offer no
insights into the behavior of the approximation as the state-relevance weights
are altered.  Therefore, it is necessary to derive new bounds for RALP which
contain $\rho$ to understand its effects.

The approach we take follows the example of a proof introduced by de Farias
and Van Roy~\yrcite{deFarias03} to bound the ALP approximation, but we extend
it match the weaker assumptions of RALP, along with the requirement that the
weights be $\lone$ regularized.

We begin by defining the relevant notation.  In the following definitions, we
will use $\Re_+$ to refer to the set of non-negative real numbers.

\begin{definition}
  \label{def:H}
  We introduce an operator $H$, defined by
  \[
  (HL)(s)=\max_{a\in\sA}\int_{\sS}p(s'|s,a)L(s')~ds',
  \]
  for all $L: \sS\rightarrow\Re_+$.
\end{definition}
Therefore, $(HL)(s)$ represents the expected value of $L$ of the \emph{next}
state if actions are chosen to maximize $L$.

\begin{definition}
  \label{def:lyapunov}
  A non-negative function $L:\sS\rightarrow \Re_+$ is a \emph{Lyapunov
  function} if there exists a subset of states $\sB$ and a $\beta_L<1$
  such that for all $s\in \sS\setminus\sB$, $\gamma(HL)(s)\leq \beta_L
  L(s)$.
\end{definition}

For an example of a Lyapunov function defined over a MDP, consider the simple
case of a random walk along the non-negative number line, so that
$s\in\ints_+$.  Assume a single action, in which the probability
\[
  p=p(s_{t+1}=\max(m-1,0)|s_t=m)>0.5
\] and
\[
p(s_{t+1}=m+1|s_t=m)=1-p.\]
If
$L(s)=s$ and $\sB=\{0\}$, then $L$ is a valid Lyapunov function, because
$L(s)$ is expected to decrease for all $s\neq 0$.

Lyapunov functions are often used to prove stability of Markov processes.
Definition \ref{def:lyapunov} differs from the definition commonly used for
stability analysis in a few ways.  First, in stability analysis, it is
required that $\sS$ be countable, and that $\sB$ be finite.  We have made
neither of these assumptions, though our bounds will be tightest when $\sB$ is
small.  The second difference is we have added a multiplicative term of
$\gamma$.  Because of these differences, a Lyapunov function as defined in
Definition \ref{def:lyapunov} may not strictly evidence stability.

Besides stability analysis, Lyapunov functions have also previously appeared
in Reinforcement Learning literature, though in different contexts from our
application~\cite{perkins03,rohanimanesh04}.

In the remainder of this section, we will occasionally refer to the weighted
max-norm, where for a vector $U$ and a function $F$,
$\norm{U}_{\infty,F}=\max_i \abs{U(s_i)\cdot F(s_i)}$, and the weighted $L_1$
norm, where $\norm{U}_{1,F}=\sum_i \abs{U(s_i)\cdot F(s_i)}$.

We will start with the following Lemma.  To conserve space, and because the
proof is similar to one presented by de Farias and Van
Roy~\yrcite{deFarias03}, we reserve the proof for Appendix \ref{app:lem1}.

\begin{lemma}
  \label{lem:defarias}
  Assume samples have been drawn from every possible state-action pair.  Let
  $\sW=\{w:\norm{w}_{-1}\leq\psi\},$ and let $w^*=\min_{w\in\sW}\norm{V^*-\Phi
  w}_\infty$.  Additionally, for a given Lyapunov function $\Phi w_L$, let
  \[
    \bar w = w^*+\lnorm{V^*-\Phi w^*}\betafrac w_L.
  \]

  If a Lyapunov function $\Phi w_L$ is constructed such that $\bar w\in\sW$,
  then,
  \[
    \norm{V^*-\Phi \tilde w}_{1,\rho}\leq \frac{2\rho^T\Phi
    w_L}{1-\beta_{\Phi w_L}} \min_{w\in\sW} \lnorm{V^*-\Phi
    w}.
  \]
\end{lemma}

We note that proving the existence of a Lyapunov function as required in the
above lemma is trivial.  First we construct a weight vector $w_L$ with all
zeros but for a positive weight corresponding to the bias feature; this
results in $\Phi w_L$ being a valid Lyapunov function.  Second, we note that
in this case $\norm{\bar w}_{-1}=\norm{w^*}_{-1}$, meeting the requirement
that $\bar w\in\sW$.

We must now remove the assumption that a sample exists for every state-action
pair.  To enable us to bound the behavior of the value function between
samples, we make the following assumption, similar to the sufficient sampling
assumption made by Petrik et al.~\yrcite{icml10}:
\begin{assumption}
  \label{ass:sufficient}
  Assume sufficient sampling, that is, for all $s\in\sS$ and $a\in\sA$, there
  exists a $\sigma\in\Sigma$ such that $\sigma^a=a$ and:
  \begin{align*}
    \norm{\phi(\sigma^s)-\phi(s)}_\infty\leq&\delta_\phi\\
    \norm{R(\sigma^s)-R(s)}_\infty\leq&\delta_R\\
    \norm{p(s'|\sigma^s,a)-p(s'|s,a)}_\infty\leq&\delta_P~\forall s'\in\sS
  \end{align*}
\end{assumption}

This assumption is not unrealistic.  For example, if the reward function,
basis functions, and transition functions are Lipschitz continuous, then
appropriate values of $\delta_\phi, \delta_R,$ and $\delta_P$ are easily
calculated given the greatest distance between any point in the state space
and a sampled point.

We describe the maximum difference between the RALP solution and the true
solution by using the limits from Assumption \ref{ass:sufficient} to
demonstrate the following Lemma:

\begin{lemma}
  \label{lem:m2}
  Let $M_1$ be an MDP with optimal value function $V^*_1$, and let $\Sigma$ be
  an incomplete set of samples drawn from $M_1$ such that not all state-action
  pairs are sampled, but Assumption \ref{ass:sufficient} is fulfilled.
  Therefore, the RALP for $M_1$ has the constraint
  $T_{\sigma^a}\Phi(\sigma^s)w\leq\Phi(\sigma^s)w$ for all $\sigma\in\Sigma$,
  and the bounded $L_1$ constraint, but is missing all other possible RALP
  constraints.

  There exists an MDP $M_2$ with an optimal value function $V^*_2$, identical
  in every way to $M_1$ but for the reward function, such that the RALP
  solution with no missing constraints is equal to the RALP solution
  constructed on $\Sigma$, and $\norm{V^*_1-V^*_2}_{\infty}\leq
  \frac{2(\delta_\phi\psi+\delta_R+\delta_P\psi)}{1-\gamma}$.
\end{lemma}
\paragraph{Proof Sketch:} We first show that if $R_1$ and $R_2$ are the
respective reward functions of $M_1$ and $M_2$,
\[
  \norm{R_1-R_2}_\infty\leq 2(\delta_\phi\psi + \delta_R+\delta_P\psi).
\]
We then show that if
$\norm{R_1-R_2}_\infty\leq \delta$,
\[
  \norm{V^*_1-V^*_2}_{\infty}\leq\frac{\delta}{1-\gamma}.
\]

We leave the details of the proof for Appendix \ref{app:lem2}.

We are now prepared to present the first result of this paper.
\begin{theorem}
  \label{thm:bound}
  Let $\Phi w_L$ be a Lyapunov function as required by Lemma
  \ref{lem:defarias}.  Define $\Sigma$, MDPs $M_1$ and $M_2$ and their
  respective optimal value functions $V^*_1$ and $V^*_2$ as in Lemma
  \ref{lem:m2}. Define $\epsilon_p=\delta_\phi\psi+\delta_R+\delta_P\psi$.
  Let $\tilde w$ be the RALP solution to $M_1$.
  \[
    \norm{V_1^*-\Phi \tilde w}_{1,\rho}\leq \frac{2\rho^T\Phi
    w_L}{1-\beta_{\Phi w_L}} \min_{w\in\sW} \lnorm{V^*_2-\Phi
    w}+\frac{2\epsilon_p}{1-\gamma}.
  \]
\end{theorem}
\paragraph{Proof:}
Because $\tilde w$ is an optimal solution given all samples from $M_2$, Lemmas
\ref{lem:defarias} and \ref{lem:m2} allow us
\[
  \norm{V_2^*-\Phi \tilde w}_{1,\rho}\leq \frac{2\rho^T\Phi
  w_L}{1-\beta_{\Phi w_L}} \min_{w\in\sW} \lnorm{V_2^*-\Phi
  w}.
\]
Additionally, Lemma \ref{lem:m2} gave us
$\norm{V_1^*-V_2^*}_\infty\leq\frac{2\epsilon_p}{1-\gamma}$.

Because $\rho$ is a probability distribution,
\[
\norm{V_1^*-V_2^*}_{1,\rho}\leq\norm{V_1^*-V_2^*}_\infty
\leq\frac{2\epsilon_p}{1-\gamma}.
\]
Due to the triangle inequality, Theorem \ref{thm:bound} follows.

\subsection{Discussion}
\label{subsec:rhoDisc}

This bound is not only tighter than those presented in previous literature,
but also allows us to analyze the RALP approximation quality in new ways.
First, we can observe which Lyapunov functions would result in a better
approximation, and discuss the characteristics of MDPs which allow for those
Lyapunov functions, and therefore lend themselves particularly well to value
function approximation by RALP.  Second, as $\rho$ now appears in our bound,
the bound provides a way of relating our choice of $\rho$ to approximation
quality, allowing for more intuitive and successful parameter assignments.  We
address these in turn.

The Lyapunov function $\Phi w_L$ appears in the first term of our bound in
three places, namely the dot product with $\rho$, the definition of
$\beta_{\Phi w_L}$, and in the norm defining the ``optimal" $w$ to which we
compare our approximation.  We first note that the bound becomes smaller as
$\beta_{\Phi w_L}$ decreases.  This suggests that the more stable the MDP, the
better RALP can approximate the value function.

This interpretation of Theorem \ref{thm:bound} leads to other intuitive
explanations.  Consider an MDP with only $L(s)=1~\forall s\in\sS$ as a
Lyapunov function.  Now assume two nearby samples, one where a ``good" action
is taken, in the direction of positive reward, and another where a ``bad"
action is taken, in the direction of negative reward.  When the linear program
is solved, due to the proximity of the two samples, the constraint
corresponding to the ``bad" sample is nearly certain to be loose, and may as
well be removed.  However, if the MDP is highly stable, then these two
extremely different samples would be unlikely, and both constraints are
candidates for being tight.  Therefore, more samples from an MDP with a small
$\beta_L$ are likely to be involved in defining the feasible space,
potentially resulting in an improved approximation.

The appearance of the Lyapunov function in the norm of the bound indicates the
bound is tighter when the feature space $\Phi$ allows for a close
approximation in areas where the Lyapunov function is small.  Bertsimas et
al.~\yrcite{bertsimas98} demonstrated that a small Lyapunov function value
correlates in expectation with a higher probability in the stationary
distribution of a Markov chain.  This is particularly interesting when
considering the appearance of the dot product between the Lyapunov function
and $\rho$.  This dot product makes it apparent that the approximation
improves when $\rho$ is large only where $\Phi w_L$ is small.  This provides
evidence that the stationary distribution of the MDP under the optimal policy
may be an advantageous setting for $\rho.$  This evidence meshes well with the
intuition that greater accuracy is most useful in frequently-visited
states.

\section{Sampling Distribution}
\label{sec:samp}

Imagine an MDP with a small finite state space and a single action.  Ideal
sampling would provide a single sample from each state, giving us an objective
function of $\sum_{s\in\sS}\rho(s)\Phi(s)w$.  However, if sampling from a
distribution across the state space, this ideal situation would be unlikely;
some states would go unsampled, while others would be sampled multiple times.
Because the objective function is defined on samples, this means states that
were sampled multiple times would appear in the objective function multiple
times, causing the linear program to tighten constraints at those states at
the expense of accuracy in other states.

Of course, a similar scenario occurs in infinite state spaces as well.
Multiple states near to each other may be sampled, while other regions have
very few samples; this encourages the linear program to choose features and
an approximate value function which tightens constraints in heavily-sampled
regions at the expense of sparsely-sampled regions.  In this section we
discuss the effects of sampling from an arbitrary distribution over the state
space $\mu$ and ways this can help the researcher understand how to design
sampling methods.

\begin{observation}
  \label{obs:sampling}
  Let $\Sigma_1$ and $\Sigma_\mu$ be sample sets of equal cardinality $N$
  drawn from the state space from the uniform distribution and an arbitrary
  distribution $\mu$, respectively.  Let $\Phi_1$ and $\Phi_\mu$ be the
  feature matrices defined over the states of sets $\Sigma_1$ and
  $\Sigma_\mu$.  For any weight vector $w$,
  \[
    \expect{\mu\Phi_1 w}=\expect{\bone\Phi_\mu w}.
  \]
\end{observation}
This observation is easy to support; both expectations equal $\int_\sS
\mu(s)\Phi(s) w~ds$.

This means that for any $w$, sampling from a non-uniform distribution $\mu$
provides an equivalent objective function in expectation to sampling from a
uniform distribution but setting the state-relevance weights equal to $\mu$.
We note this is not equivalent to expecting the same approximate value
function as the constraints remain different; this makes the effect of
altering the sampling distribution greater than that of altering the objective
function alone.

Additionally, the bound presented in Theorem \ref{thm:bound} offers an
interpretation of results from sampling from a distribution.  As sampling
becomes less uniform, $\epsilon_p$ likely increases, due to the existence of
larger unsampled regions.  For example, this would happen in the
Lipschitz-continuous case for fulfilling Assumption \ref{ass:sufficient}.
However, for a sampling distribution which is dense where the Lyapunov value
is small, then the total Lyapunov value in the numerator of the first addend
is small, as well.  Therefore, it may be that the ideal sampling distribution
is one which is dense where the Lyapunov value is low, but still provides
sufficient coverage for $\epsilon_p$ to be small.

\section{Experimental Results}
\label{sec:experiments}

In this section, we demonstrate experimentally the conclusions drawn in the
previous sections.  Previous literature has already demonstrated RALP's
effectiveness in common benchmark domains; the purpose of this section
therefore, is to clearly illustrate the conclusions of the previous sections.
In a simple, easily visualized domain, we make a series of comparisons.
First, we compare the approximation accuracy of sampling from a domain with a
stable Lyapunov function to the accuracy resulting from sampling from a domain
without such a function.  Next, we compare the accuracy of the approximation
resulting from sampling uniformly to the accuracy of the approximation
resulting from sampling from two different nonuniform distributions.  Finally,
we compare the approximation accuracy of calculating an approximation with
$\rho=\bone$ to the approximation accuracy of calculating an approximation
when $\rho$ is nonuniform.  This is demonstrated using two different,
nonuniform distributions.

\begin{figure*}
\centering
\begin{tabular}{|c|c|c|}
  \hline
  \subfigure[Average error from sampling from stable domain - Average error
  from sampling from unstable domain\label{subfig:lyapunov}]{
    \includegraphics[width=2.15in]{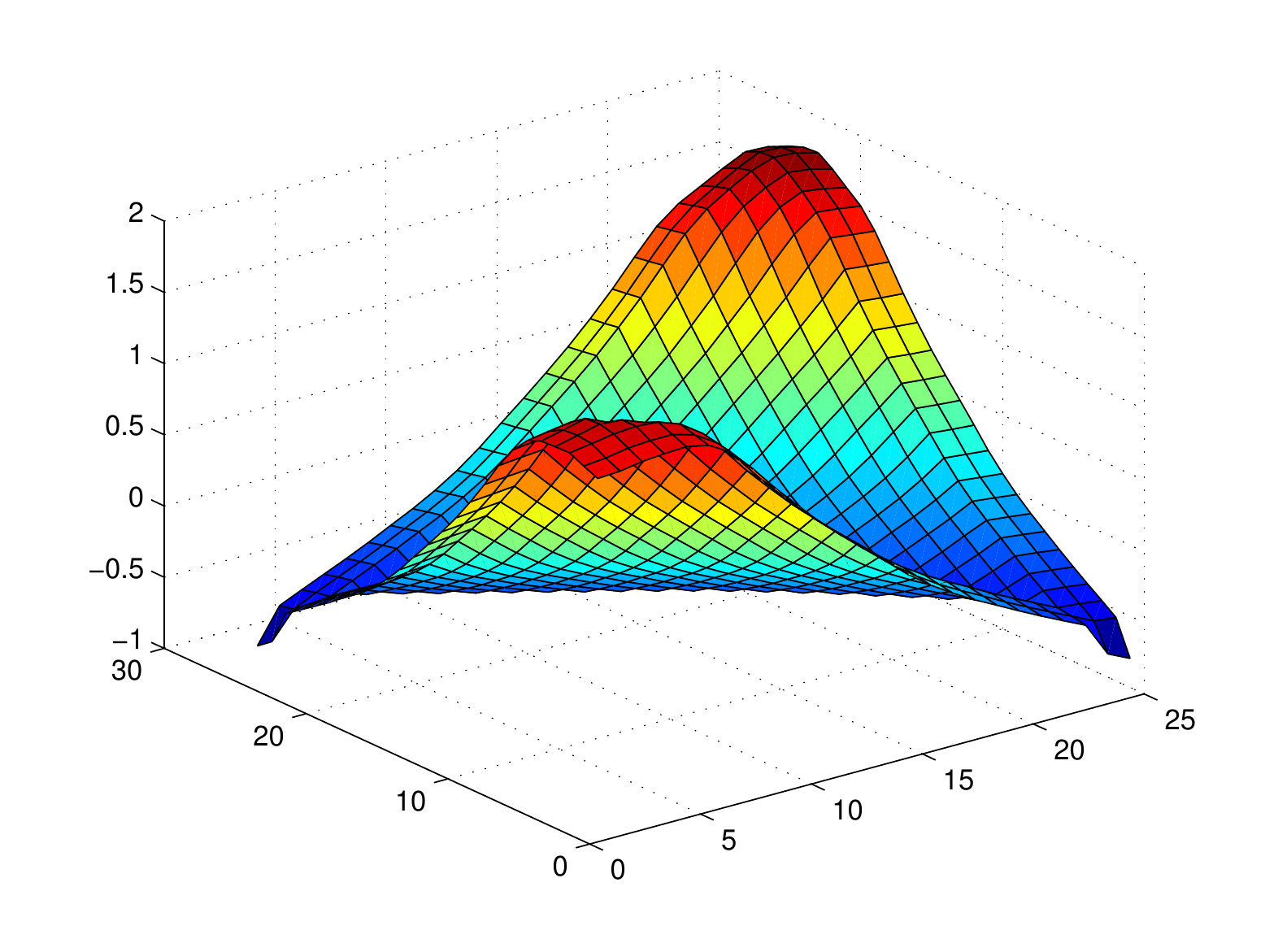}}
    &
  \subfigure[Average error from sampling from $\bone$ - Average error from
  sampling from $\zeta$\label{subfig:samp}]{
  \includegraphics[width=2.15in]{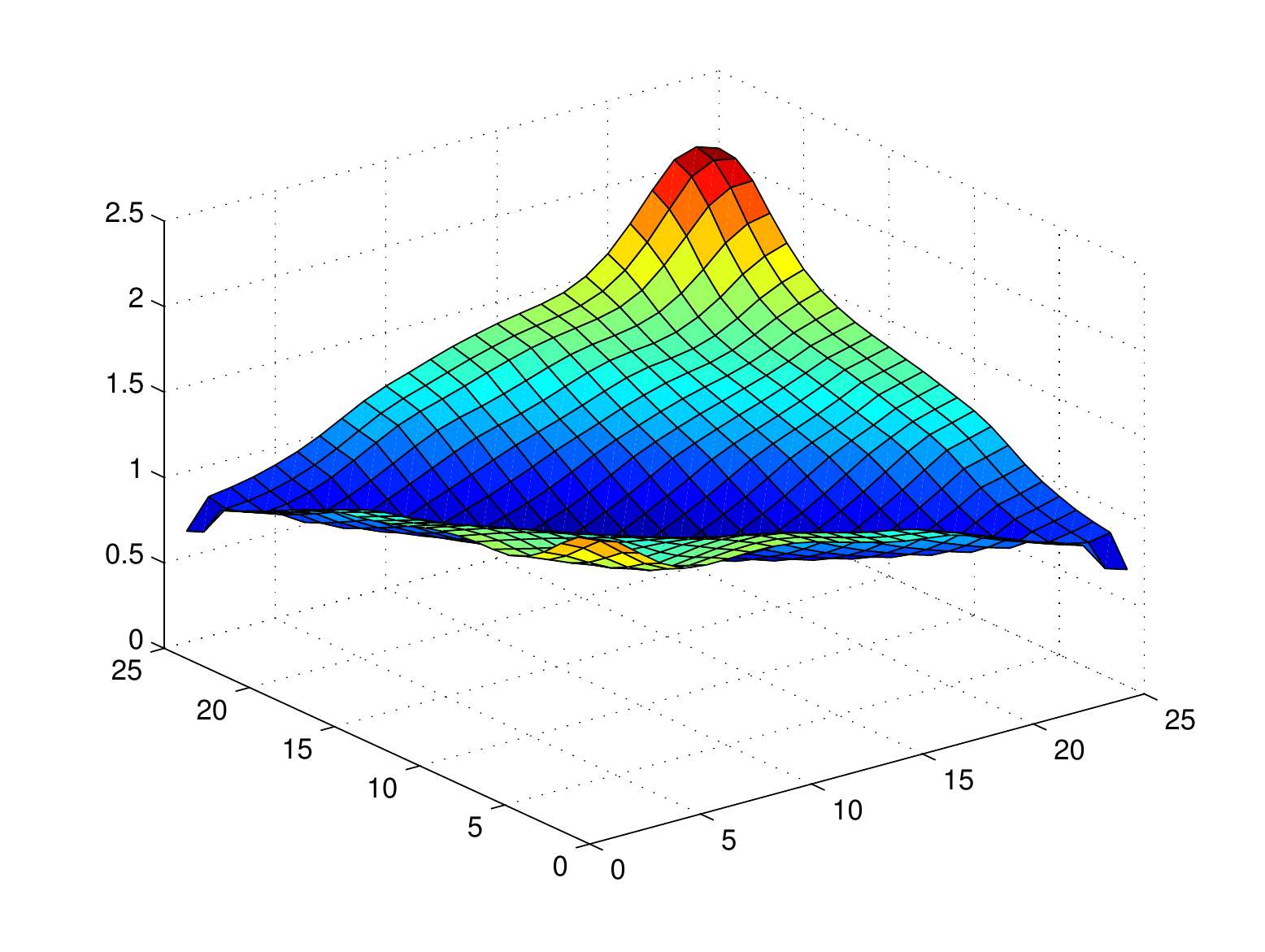}}
    &
  \subfigure[Average error from $\rho=\bone$ - Average error from
  $\rho=\zeta$\label{subfig:rho}]{ \includegraphics[width=2.15in]{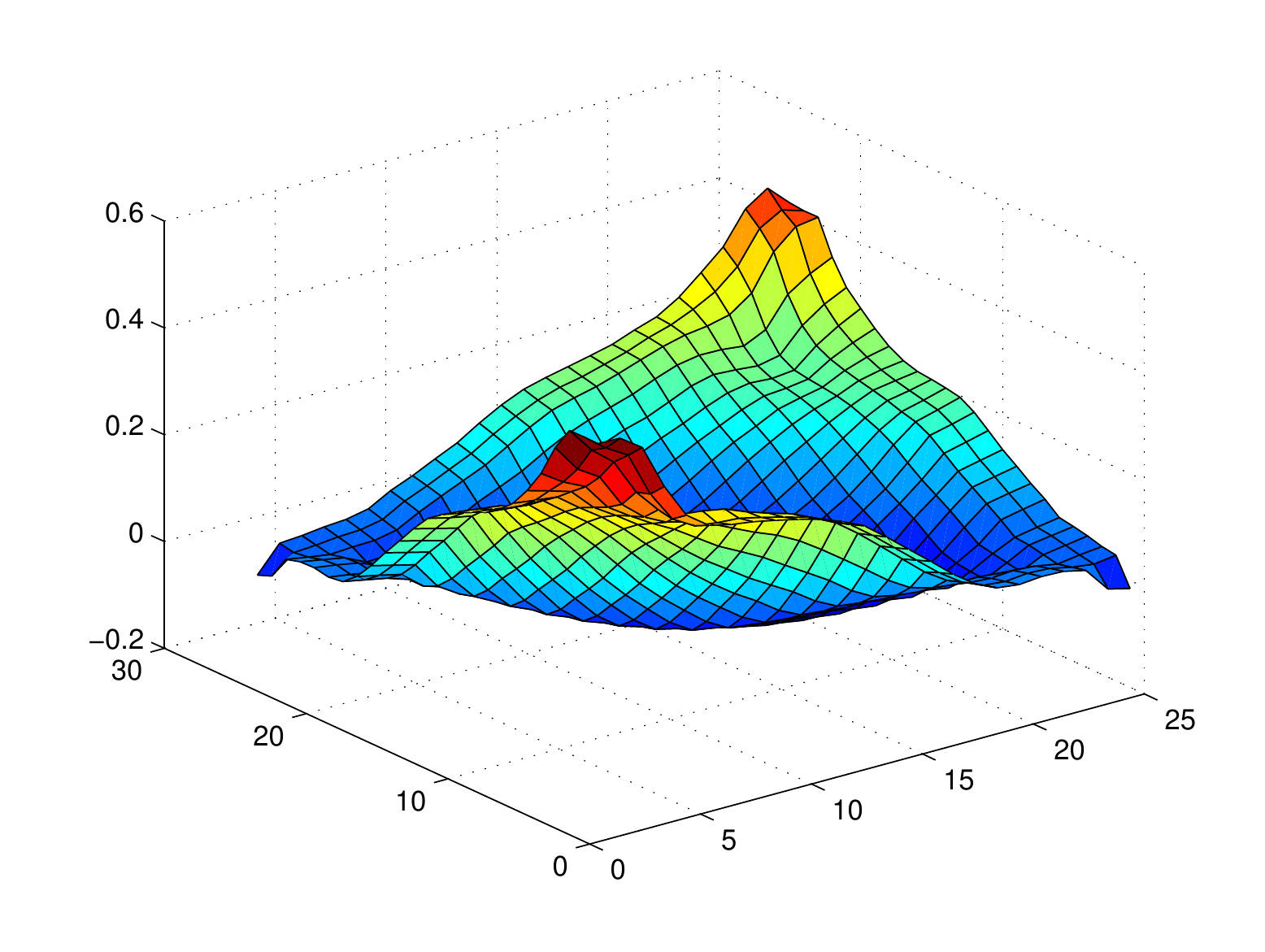}} \\
    &
  \subfigure[Average error from sampling from $\bone$ - Average error from
  sampling from $(\bone-\zeta)$\label{subfig:sampInvDiff}]{
    \includegraphics[width=2.15in]{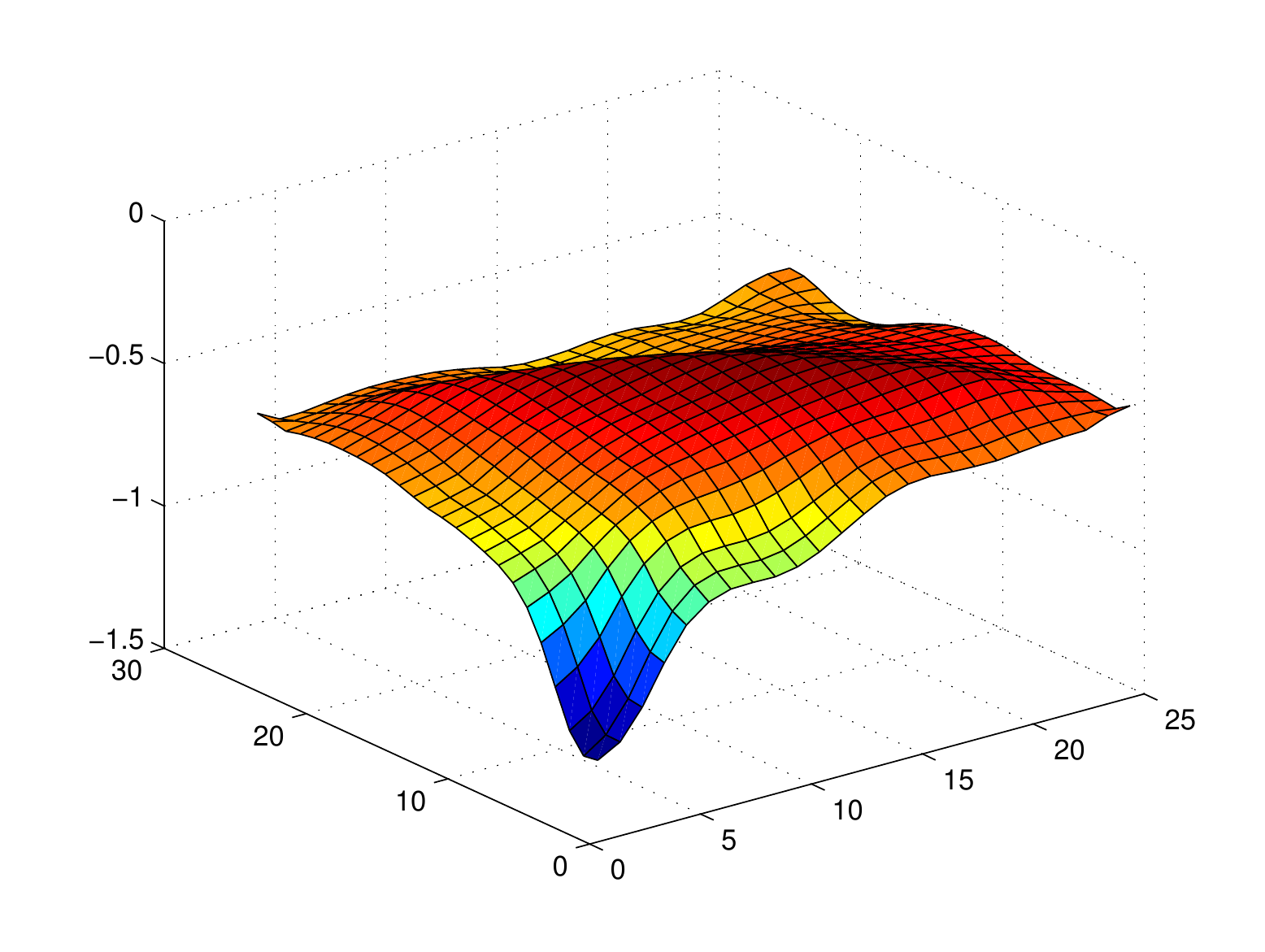}}
    &
  \subfigure[Average error from $\rho=\bone$ - Average error from
  $\rho=(\bone-\zeta)$\label{subfig:rhoInv}]{
  \includegraphics[width=2.15in]{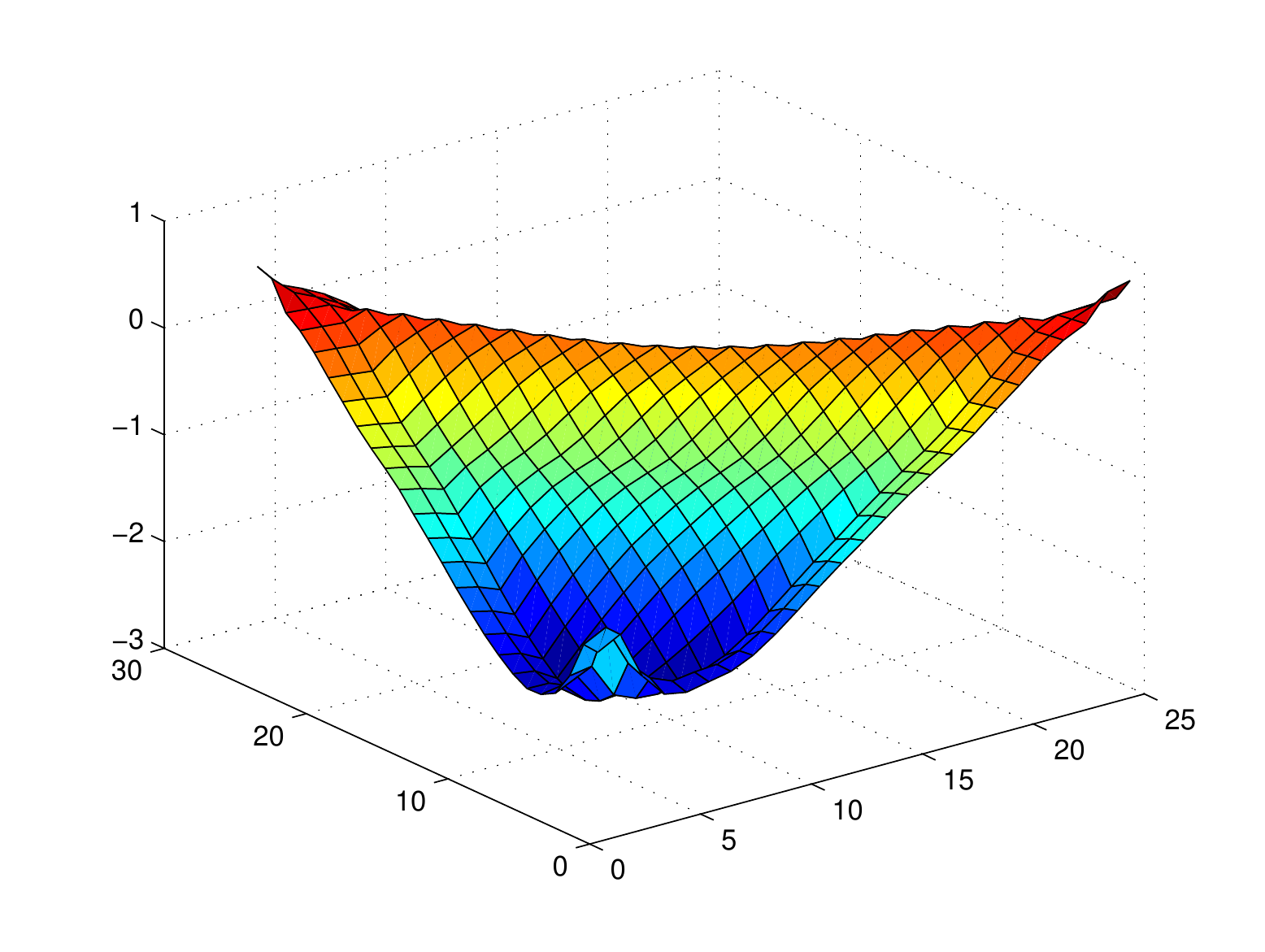}}\\
    \hline
\end{tabular}
\caption{Difference in Average Error}
\label{fig:results}
\end{figure*}

The results were obtained by drawing samples and calculating an approximation
500 times for each compared approach; the error $\abs{V^*-\Phi w}$
was then calculated, and averaged across all 500 trials.  Finally, we
calculate and display the difference between the average errors from the two
approaches.  So, if $\hat V_i^A$ is the approximation from the $i$-th run on
approach A, then when comparing two approaches, A and B, a point on the graphs
of Figure \ref{fig:results} equals
\[
  \sum_{i=1}^{500}\frac{\abs{V^*(s)-\hat
  V_i^A(s)}}{500}-\sum_{i=1}^{500}\frac{\abs{V^*(s)-\hat V_i^B(s)}}{500}.
\]

\subsection{Domain}

\begin{figure}
\centering
\mbox{
  \subfigure[Reward Regions\label{subfig:domain}]{
    \includegraphics[width=1in]{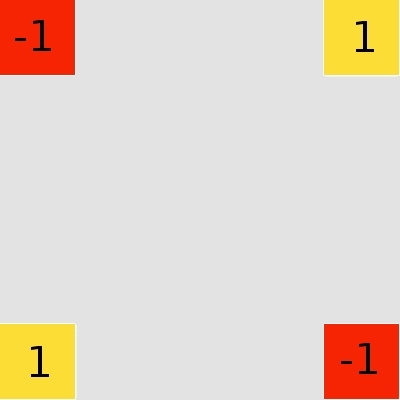}}\quad
  \subfigure[$V^*$\label{subfig:Vstar}]{
    \includegraphics[width=1.5in]{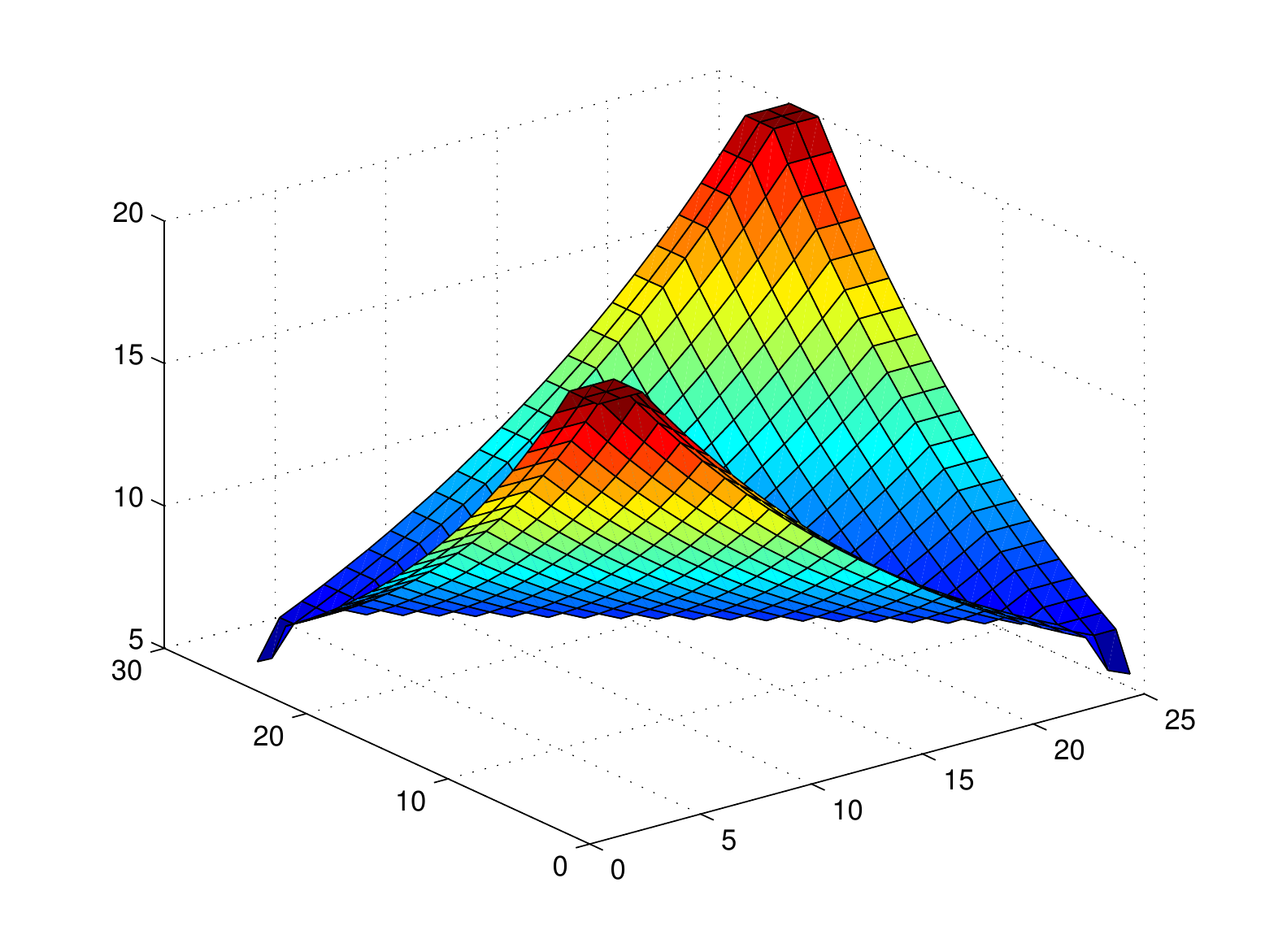}}
  }
\caption{Room Domain}
\label{fig:domain}
\end{figure}

All of the experiments were run on a domain defined by a 25 by 25 grid world.
This world included four reward regions in the corners of the grid, each of
which consisted of 9 states, as can be seen in Figure \ref{subfig:domain}.
Two reward regions (colored gold) had a reward of 1, while the others had a
reward of -1 (colored red).  The remaining states had a reward of 0. Actions
were to move one square in any of the four directions, unless constrained by a
wall, in which case that action would result in no movement.  The discount
factor $\gamma$ was set to 0.95.  For all trials, the feature set consisted of
symmetric Gaussian features centered around each $\sigma^s$ with variances of
2, 5, 10, 15, 25, 50, and 75, plus the bias feature, resulting in $9n+1$
features for $n$ samples.  The optimal value function can be seen in Figure
\ref{subfig:Vstar}.

This value function is an easy one for RALP to approximate with proper
settings of the regularization parameter and sufficient sampling.  The number
of samples and choices of the regularization parameter $\psi$ were therefore
chosen to illustrate the differences between the results of the compared
methods, not to optimize performance.

\subsection{Lyapunov Stable Domain}
\label{subsec:lyapunov}

First, we demonstrate the improvement in RALP's approximation when the domain
has a stable Lyapunov function.  A stable Lyapunov function can be created by
forcing the actor into a defined area in the state space.  In order to keep
the representational difficulty of the optimal value functions the same, we
created a Lyapunov function by eliminating actions which move the actor
further from the nearest positive reward.  This preserves the optimal policy
of the unaltered domain, keeping the optimal value functions identical, making
approximation accuracy a fair comparison. 

In the domain without a stable Lyapunov function, the actor was free to move
in the state space based on a random choice among the four actions. However,
in the domain with a stable Lyapunov function, the actor was only allowed to
move in the two directions which would not move it further from the nearest
goal.  We note that not all remaining actions are optimal, so sampling still
includes off-policy samples.

This creates a Lyapunov function where $L(s)$ equals the Manhattan distance
from $s$ to the nearest of state (1,1) or (25,25), and
$\sB=\{(1,1),(25,25)\}$.  In each trial, we uniformly sampled 20 samples.  The
regularization parameter $\psi$ was set to 0.2 and $\rho$ was $\bone$.

The result of subtracting the average errors of the approximation from the
domain with a Lyapunov function from the domain without a Lyapunov function
can be seen in Figure \ref{subfig:lyapunov}.  Therefore, positive values
indicate higher error from the domain without a Lyapunov function.

The results show that an approximation learned from samples drawn from a
stable domain is more accurate than an approximation learned from a less
stable domain everywhere except for where $s$ is near equal distance from the
two goal states.  This reinforces the intuition from Subsection
\ref{subsec:rhoDisc} that samples drawn from a
stable domain are more effective than those that are not, particularly near
the most heavily-visited regions.

\subsection{Sampling from a Nonuniform Distribution}
\label{subsec:sampling}

Next we illustrate the change in approximation accuracy when sampling from a
nonuniform distribution $\mu$.  Section \ref{sec:samp} presents evidence that
a distribution which is most dense where $L(s)$ is smallest may be
advantageous.  To create such a distribution, an agent was started at a random
state, and was allowed to take the optimal policy for 25 steps.  This was done
10,000 times, and the number of visits to each state was tabulated and
normalized.  This defined our distribution, which was heaviest on the edges
and reward corners, and otherwise slightly increasing with increasing
proximity to the positive reward regions.  We will refer to this distribution
as $\zeta$.

For these trials, 20 samples were drawn per run from the domain with the
stable Lyapunov function as discussed in Subsection \ref{subsec:lyapunov}.
The regularization parameter $\psi$ was set to 1.5, and $\rho=\bone$.  The
average error from the 500 uniformly sampled runs was subtracted from the
average error from the 500 runs with $\mu=\zeta$; the result can be seen
in Figure \ref{subfig:samp}.  Because the error from the nonuniform sampling
was subtracted from the error from uniform sampling, the positive difference
indicates the results from sampling from $\zeta$ were superior.  The results show the distribution met the goal from Section \ref{sec:samp};
sampling was varied enough to keep $\epsilon_p$ low, while dense enough in the
areas where $L(s)$ was small.

We then subtracted $\zeta$ from $\bone$ and normalized, making a distribution
we will refer to as $\bone-\zeta$, which was largest where an agent was least
likely to traverse in the stable domain.  We subtracted the average error from
sampling from $\mu=\bone-\zeta$ from the average error from sampling
uniformly, to produce Figure \ref{subfig:sampInvDiff}.  A positive value would
indicate larger error from the approximations on uniformly sampled states.
However, through the entirety of the state space, and particularly in the most
trafficked areas, sampling from $\bone-\zeta$ gave us an inferior result.

This provides evidence for the conclusions of Section \ref{sec:samp} that a
sampling distribution which is densest in the areas where the Lyapunov
function is smallest would produce the best approximations.

\subsection{Changing the State-Relevance Weights}

Lastly, we illustrate the effect on the approximation of changing the
state-relevance weights.  200 samples were drawn uniformly from the state
space; whereas the effects from the previous two experiments are most
pronounced when samples are sparse, the effects from altering $\rho$ are most
pronounced when a number of constraints can be tightened in a given region.

We would prefer to set $\rho$ to the stationary distribution; however, because
the domain is not recurrent, this is not an option.  However, the distribution
$\zeta$ created for sampling in Subsection \ref{subsec:sampling} is large
where the Lyapunov function is small, making it a reasonable replacement.  In
one set of trials, $\rho(s)$ was set to the value of this distribution at that
state; in the other, $\rho=\bone$.  $\psi$ was set to 4.  Average error from
the approximation resulting from a nonuniform $\rho$ was subtracted from
average error from the approximation resulting from a uniform $\rho$.
Therefore, a positive value indicates a better approximation from the
nonuniform $\rho.$ Sampling was done uniformly from the stable domain.

Figure \ref{subfig:rho} shows that nearly the entire state space was more
accurate with a nonuniform $\rho$, except for where $\zeta(s)\approx 0$.  The
difference is small because with a large number of samples, both
approximations were quite accurate. 

In addition, we compared the uniform $\rho$ approximation to an approximation
using $\rho=\bone-\zeta$, which is large where the Lyapunov value is large,
resulting in an increased dot product $\rho\Phi w$ in Theorem \ref{thm:bound}.
Again, we subtracted the error of the approximation resulting from
$\rho=\zeta-1$ from the error of the approximation resulting from
$\rho=\bone$, producing Figure \ref{subfig:rhoInv}.  A positive value
indicates the nonuniform $\rho$ approximated that state better than did using
a uniform $\rho$.  However, there are few positive values as the use of
$\rho=\bone-\zeta$ resulted in a dramatically inferior approximation, particularly
in areas where $\rho$ was small.

From both figures, it is clear that a higher $\rho$ value in a given portion
of the state space resulted in an improved approximation, particularly if
designed with Theorem \ref{thm:bound} in mind.

\section{Conclusion}
\label{sec:conclusion}

The experimental success of RALP in previous literature, along with its
easily-fulfilled assumptions, suggests promise for its application to
real-life, complicated, and complex domains.  Despite this promise, and
despite the evidence that the effects are dramatic, no theory had been
produced to analyze changes in the approximation quality given changes to the
objective function parameter $\rho,$ or due to differences in sampling
strategies.  These considerations are essential to the use of RALP in the real
world; it is rarely possible to sample uniformly, and the importance of
accuracy across the state space is rarely consistent.

In this paper, we demonstrate the importance of understanding these ideas, and
produce a bound on the approximation error of RALP which is tighter and more
informative than previous bounds.  This bound provides intuition into the
quality of the RALP approximation as a function of state-relevance weights and
sampling distributions.  In addition, we demonstrated that the quality of a
RALP approximation is particularly good when the domain is stable and has a
Lyapunov function with a small $\beta_L$.

Future work remains, particularly in the area of solving the linear program
quickly in the presence of large amount of data.  Though convex optimization
solvers are considered ``fast," with large amounts of data, the memory and
time requirements may be too large for realistic use.  Fortunately, it may be
that the structure of the problem, the small percentage of tight constraints,
and small percentage of active features will avail itself to faster, but
equivalent, approaches.

\section*{Acknowledgements}
Thank you to the anonymous reviewers for their help in improving this paper.
Additionally, we are grateful for support from the Naval Research Laboratory
Information Management \& Decision Architecture Branch (Code 5580), as well as
financial support by the Office of Naval Research, grant numbers
N001613WX20992 and N0001414WX20507.

\bibliography{gbib}
\bibliographystyle{icml2014}
\appendix
\section{Proof of Lemma 1}
\label{app:lem1}
As stated in Section 4, this proof is very similar to Theorem 3 by de Farias
and van Roy (2003), but we include it for clarity nonetheless.  The Lemma
constructs a point in the feasible space of the linear program which provides
an approximation with bounded error, and then shows the point chosen by RALP
must be no further than that constructed point.  The proof first requires a
series of additional Lemmas.

\begin{lemma}
\label{lem:herthree}
For any functions $V$ and $\bar V$,
\[
\abs{T\bar V-TV}\leq\gamma \max_\pi P_\pi \abs{\bar V-V}.
\]
\end{lemma}
\begin{proof} For any $V$ and $\bar V$,
\begin{align*}
T\bar V - TV =& \max_\pi(R+\gamma P_\pi \bar V)-\max_\pi(R+\gamma P_\pi V)\\
=&R+\gamma P_{\pi_{\bar V}} \bar V-R-\gamma P_{\pi_V} V \\
\leq&\gamma \max_\pi P_\pi (\bar V - V)\\
\leq&\gamma \max_\pi P_\pi \abs{\bar V - V},
\end{align*}
where $\pi_V$ and $\pi_{\bar V}$ represent the greedy policies with respect to
value functions $V$ and $\bar V$.  By reversing the terms, we can show
$TV-T\bar V\leq\gamma \max_\pi P_\pi \abs{\bar V - V}$, leading to our result.
\end{proof}

\begin{lemma}
\label{lem:herfour}
For any vector $L$ with positive components and any vector $V$,
\[
TV\leq V+(\gamma HL+L)\|V-V^*\|_{\infty,\frac{1}{L}}.
\]
\end{lemma}
\begin{proof} Note that
\[
\abs{V^*(s)-V(s)}\leq\|V-V^*\|_{\infty,\frac{1}{L}}V(s).
\]

Because of Lemma \ref{lem:herthree},
\begin{align*}
|(TV)(s)-(TV^*)(s)|
\leq&\gamma\max_\pi\sum_{s'\in\sS}P_\pi(s,s')\abs{V(s')-V^*(s')}\\
\leq&\gamma\|V-V^*\|_{\infty,\frac{1}{L}}\max_{a\in\sA}\sum_{s'\in\sS}P_a(s,s')L(s')\\
   =&\gamma\|V-V^*\|_{\infty,\frac{1}{L}}(HL)(s).
\end{align*}

Define $\epsilon=\|V-V^*\|_{\infty,\frac{1}{L}}$.
\begin{align*}
  TV(s)\leq& V^*(s)+\gamma\epsilon(HL)(s)\\
  \leq& V(s)+\epsilon L(s)+\gamma\epsilon(HL)(s).
\end{align*}
\end{proof}

\begin{lemma}
\label{lem:herfive}
Let $w_L$ be a weight vector such that $\Phi w_L$ is a Lyapunov function, $w$
be an arbitrary weight vector, and
\[
\bar w = w+\lnorm{V^*-\Phi w}\betafrac w_L.
\]
Then, $T\Phi\bar w\leq \Phi \bar w$.
\end{lemma}
\begin{proof}
Let $\epsilon=\lnorm{V^*-\Phi w}$.  For any state
$s\in\sS$,
\begin{small}
\begin{align*}
|(T\Phi&\bar w)(s)-(T\Phi w)(s)|\\
=&\abs{\left(T\left[(\Phi w + \epsilon\betafrac \Phi
w_L\right]\right)(s)-(T\Phi w)(s)}\\
\leq&\gamma\max_\pi \sum_{s'\in\sS}P_\pi(s,s')\\
    &\cdot\abs{(\Phi w(s') +
\epsilon\betafrac \Phi w_L(s'))-\Phi w(s')}\\
\leq&\gamma\max_\pi \sum_{s'\in\sS}P_\pi(s,s')\epsilon\betafrac(\Phi
w_L)(s')\\
=&\gamma\epsilon\betafrac(H\Phi w_L)(s).
\end{align*}
\end{small}
The first line is a replacement of $\bar w$ with its definition, the second is
due to Lemma \ref{lem:herthree}, the third is due to the cancellation of the
two $\Phi w$ terms and the fact that because $\Phi w_L$ is a Lyapunov
function, $\betafrac>0$ (note this is true for states in sets $\sB$ and
$\sS\setminus\sB$ as defined in Definition 1), and the final line is due to
Definition 2.

From this, we can conclude
\[
T\Phi\bar w\leq T\Phi w + \gamma \epsilon\betafrac H\Phi w_L.
\]

We can apply Lemma \ref{lem:herfour} to get
\[
T\Phi \bar w \leq \Phi w + \epsilon(\gamma H\Phi w_L+\Phi w_L),
\]
and therefore,
\begin{small}
\begin{align*}
T\Phi\bar w \leq& \Phi w + \epsilon(\gamma H\Phi w_L+\Phi w_L) +
\gamma\epsilon\betafrac H\Phi w_L\\
=&\Phi w+\epsilon\betafrac \Phi w_L-\epsilon\betafrac \Phi w_L\\
 &+ \epsilon(\gamma H\Phi w_L+\Phi w_L)+ \gamma\epsilon\betafrac H\Phi w_L\\
=&\Phi \bar w-\epsilon\betafrac \Phi w_L+\epsilon(\gamma H\Phi w_L+\Phi
w_L)\\
&+ \gamma\epsilon\betafrac H\Phi w_L\\
=&\Phi \bar w+ \epsilon(\gamma H\Phi w_L+\Phi w_L)- \epsilon\betafrac(\Phi w_L-\gamma H\Phi w_L)\\
\leq&\Phi \bar w+ \epsilon(\gamma H\Phi w_L+\Phi w_L)-\epsilon(\Phi w_L+\gamma H\Phi w_L)\\
=&\Phi \bar w.
\end{align*}
\end{small}

The penultimate line can be shown given that $\Phi w_L-\gamma H\Phi w_L >0$
and
\begin{small}
\begin{align*}
\frac{2}{1-\beta_{\Phi
w_L}}-1=&\frac{2}{1-\max_{s\in\sS\setminus\sB}((\gamma(H\Phi w_L)(s))/((\Phi
w_L)(s)))}-1\\
=&\max_{s\in\sS\setminus\sB}\frac{(\Phi w_L)(s)+\gamma(H\Phi w_L)(s)}{(\Phi w_L)(s)-\gamma(H\Phi w_L)(s)}
\end{align*}
\end{small}
\end{proof}

Lemma \ref{lem:herfive} demonstrates that all constraints in RALP will be
satisfied by $\bar w$, with the exception of the constraint enforcing the
$L_1$ regularization.  However, we have required even this constraint to be
satisfied by requiring $\bar w\in\sW$.  Therefore, $\bar w$ lies in the
feasible region for RALP.

\begin{lemma}
\label{lem:bigger}
If every state-action pair is represented with a constraint in the RALP, a
vector $\tilde w$ solves the RALP if and only if it solves
\begin{align*}
\argmin_w \hspace{.3cm}&\norm{V^*-\Phi w}_{1,\rho}\\
s.t. \hspace{.5cm}&T_a\Phi(s) w\leq\Phi(s) w
~~\forall s\in\sS, a\in\sA\\
\hspace{.5cm}&\norm{w_{-1}}_1\leq \psi
\end{align*}
\end{lemma}

\begin{proof}
For any policy $\pi$, the Bellman operator $T\pi$ is a contraction in max
norm.  If the Bellman error is one-sided, $T$ is also monotonic.  Therefore,
for any $V$ such that $V\geq TV$,
\[
V\geq TV \geq T^2V \geq V^*.
\]
Therefore, any $w$ that is a feasible solution to a RALP satisfies
$\Phi w\geq V^*$.  From this, we can conclude
\begin{align*}
\norm{V^*-\Phi w}_{1,\rho}=&\sum_{x\in S}\rho(x)\abs{V^*(x)-\Phi(x) w}\\
=&\rho^T\Phi w-\rho^T V^*.
\end{align*}
Because $V^*$ is constant, minimizing $\rho^T\Phi w$ with RALP
constraints is equivalent to minimizing $\norm{V^*-\Phi w}_{1,\rho}$
with RALP constraints.
\end{proof}

Given Lemmas \ref{lem:herfive} and \ref{lem:bigger}, we can finally prove
Lemma 1.
\begin{align*}
  \norm{V^*-\Phi \tilde w}_{1,\rho}\leq& \norm{V^*-\Phi\bar w}_{1,\rho}\\
           =&\sum_{s\in\sS}\rho(s)\abs{V^*-(\Phi\bar w)(s)}\\
           =&\sum_{s\in\sS}\rho(s)(\Phi w_L)(s)\frac{\abs{V^*-(\Phi\bar
w)(s)}}{(\Phi w_L)(s)}\\
\leq&\left(\sum_{s\in\sS}\rho(s)(\Phi w_L)(s)\right)
\max_{s'\in\sS}\frac{\abs{V^*-(\Phi\bar w)(s')}}{(\Phi w_L)(s')}\\
=&\rho^T\Phi w_L\norm{V^*-\Phi\bar w}_{\infty,1/\Phi w_L}\\
\leq&\rho^T\Phi w_L\left(\norm{V^*-\Phi w^*}_{\infty,1/\Phi
w_L}+\norm{\Phi\bar w-\Phi w^*}_{\infty,1/\Phi w_L}\right)\\
\leq&\rho^T\Phi w_L (\norm{V^*-\Phi w^*}_{\infty,1/\Phi
w_L}\\
&+\norm{V^*-\Phi w^*}_{\infty,1/\Phi w_L} \left(\frac{2}{1-\beta_{\Phi
w_L}}-1\right)\norm{\Phi w_L}_{\infty,1/\Phi w_L})\\
=&\frac{2\rho^T\Phi w_L}{1-\beta_{\Phi w_L}}\norm{V^*-\Phi
w^*}_{\infty,1/\Phi w_L}.
\end{align*}
The penultimate line is due to the definition of $\bar w$, and the final line
occurs because $\norm{\Phi w_L}_{\infty,1/\Phi w_L}=1$.\hfill$\blacksquare$

\section{Proof of Lemma 2}
\label{app:lem2}

When a constraint does not exist in RALP for some state, this does not mean
the value at that state is completely unconstrained; because we bounded the
rate of change of all components of the approximate and true value functions
in Assumption 1, the existence of a constraint constructed on a nearby state
means the existence of what we will call an \emph{implied} constraint.

This lemma explicitly constructs these implied constraints, and quantifies the
maximum distance from the true constraint which would have existed had that
state been sampled.  It does this by building two MDPs, identical in every
way, but for the reward function.  $M_1$ has been incompletely sampled, with
sample set $\Sigma$.  Every state-action pair therefore has either an explicit
or implied constraint in the corresponding RALP.  $M_2$, however, has been
completely sampled.  Every state-action pair in the set $\Sigma$ is identical
in $M_2$, but all state-action pairs not in $\Sigma$ have a sample producing a
constraint identical to the implied constraints of $M_1$.  Because the
constraints are the same, the RALP solution is the same.  We demonstrate the
difference in the reward functions $R_1$ and $R_2$ is bounded, and thus, the
difference in the optimal value functions $V^*_1$ and $V^*_2$ is bounded.

\begin{lemma}
\label{lem:reward}
Given an MDP $M_1$ such that Assumption 1 is true and incomplete sample set
$\Sigma$, an MDP $M_2$ exists such that constructing the RALP with constraints
for all state-action pairs results in an identical RALP solution to that of
the RALP constructed from $\Sigma$, and
$\norm{R_1-R_2}_\infty\leq2(\delta_\phi\psi+\delta_R+\delta_P\psi)$.
\end{lemma}

\begin{proof}
Consider an arbitrary state-action pair $s,a$, which is not represented by a
sample in $\Sigma$.  This means we are missing the constraint
\begin{equation}
\label{eqn:wish}
R_1(s)+\gamma \sum_{x\in\sS}\left[p(x|s,a)\Phi(x)\right] w \leq
\Phi(s) w.
\end{equation}
Let us refer to the sample in $\Sigma$ which fulfills the sampling assumption
with $s$ and $a$ as $\sigma$.  We can now construct a bound for how incorrect
each component of this constraint can be if we use the constraint at $\sigma$
and our sampling assumption to replace the missing constraint.  For instance,
the reward function $R(s)$ is easily bounded.
\begin{equation*}
\label{eqn:rewbound}
R_1(\sigma^s)-\delta_R\leq R_1(s) \leq R_1(\sigma^s)+\delta_R
\end{equation*}
We now bound $\Phi(s) w$.  Because the sampling assumption allows each basis
function to change only a finite amount, and because $\norm{ w_{-1}}_1\leq
\psi$, and $\bone(s)=\bone(\sigma^s)$,
\begin{equation*}
\label{eqn:phibound}
\Phi(\sigma^s) w-\delta_\Phi\psi
\leq\Phi(s) w\leq\Phi(\sigma^s) w+\delta_\Phi \psi
\end{equation*}
The final component is $\gamma \sum_{x\in \sS}p(x|s,a)\Phi(x) w$,
which expresses our expected value at the next state.  It will be convenient
to separate the bias feature $\bone$ from the rest of $\Phi$.  We will denote
the remainder of the design matrix as $\Phi_{-1}$, and the weights that
correspond to $\Phi_{-1}$ as $w_{-1}$.  Similarly, we will denote the weight
corresponding to $\bone$ as $w_1$.
\begin{align*}
\sum_{x\in \sS}p(x|s,a)\Phi(x) w=&
\sum_{x\in\sS}p(x|s,a) w_1
+ \sum_{x\in\sS}p(x|s,a)\Phi_{-1}(x) w_{-1}\\
=&  w_1+\sum_{x\in\sS}p(x|s,a)
\Phi_{-1}(x) w_{-1}
\end{align*}
Again, we have bounded the allowable change in our expression of probability.
\begin{align*}
 w_1+&\sum_{x\in\sS}p(x|s,a) \Phi_{-1}(x) w_{-1}\\
\leq& w_1+\sum_{x\in\sS}\left[p(x|\sigma^s,\sigma^a)+
\delta_P\right]\Phi_{-1}(x) w_{-1}\\ =&
w_1+\sum_{x\in\sS}p(x|\sigma^s,\sigma^a)\Phi_{-1}(x) w_{-1}+
\delta_P\sum_{x\in\sS}\Phi_{-1}(x) w_{-1}
\end{align*}
Because each basis function $\Phi$ is can be standardized such that
$\norm{\Phi}_1=1$, and because $\norm{w_{-1}}_1\leq \psi$, the second
summation can be at most $\psi$.  So,
\begin{align*}
\label{eqn:pbound}
\sum_{x\in\sS}p(x|\sigma^s,\sigma^a)\Phi(x) w-\delta_P \psi&\leq
\sum_{x\in\sS}\left[p(x|s,a)\Phi(x)\right] w\\
&\leq \sum_{x\in\sS}p(x|\sigma^s,\sigma^a)\Phi(x) w+\delta_P\psi.
\end{align*}

We now combine these results, and construct our implied
constraint to take the place of the missing constraint expressed by Equation
\ref{eqn:wish}.  We see that the maximum possible change by the
approximate value function is $\delta_\Phi \psi + \delta_R + \delta_P \psi$.
So, the total cumulative error in the constraint is at most $2(\delta_\Phi
\psi + \delta_R + \delta_P \psi)$.  So, we effectively have the following
constraint:
\begin{equation*}
\label{eqn:qer}
R_1(s)+q- \gamma \sum_{x\in\sS}\left[p(x|s,a) \Phi(x)\right] w \geq
\Phi(s) w,
\end{equation*}
where $\abs{q}\leq 2(\delta_\Phi \psi + \delta_R + \delta_P \psi)$.

Let $M_2$ be an MDP which is identical in every way to $M_1$, except
$R_2(s)=R_1(s)+q$.  The RALP solution for $M_1$ will be equivalent to the RALP
solution for $M_2$, and $\norm{R_1-R_2}_\infty\leq
2(\delta_\Phi\psi+\delta_R+\delta_P\psi).$
\end{proof}

\begin{figure}[h]
\begin{center}
\includegraphics[width=2in]{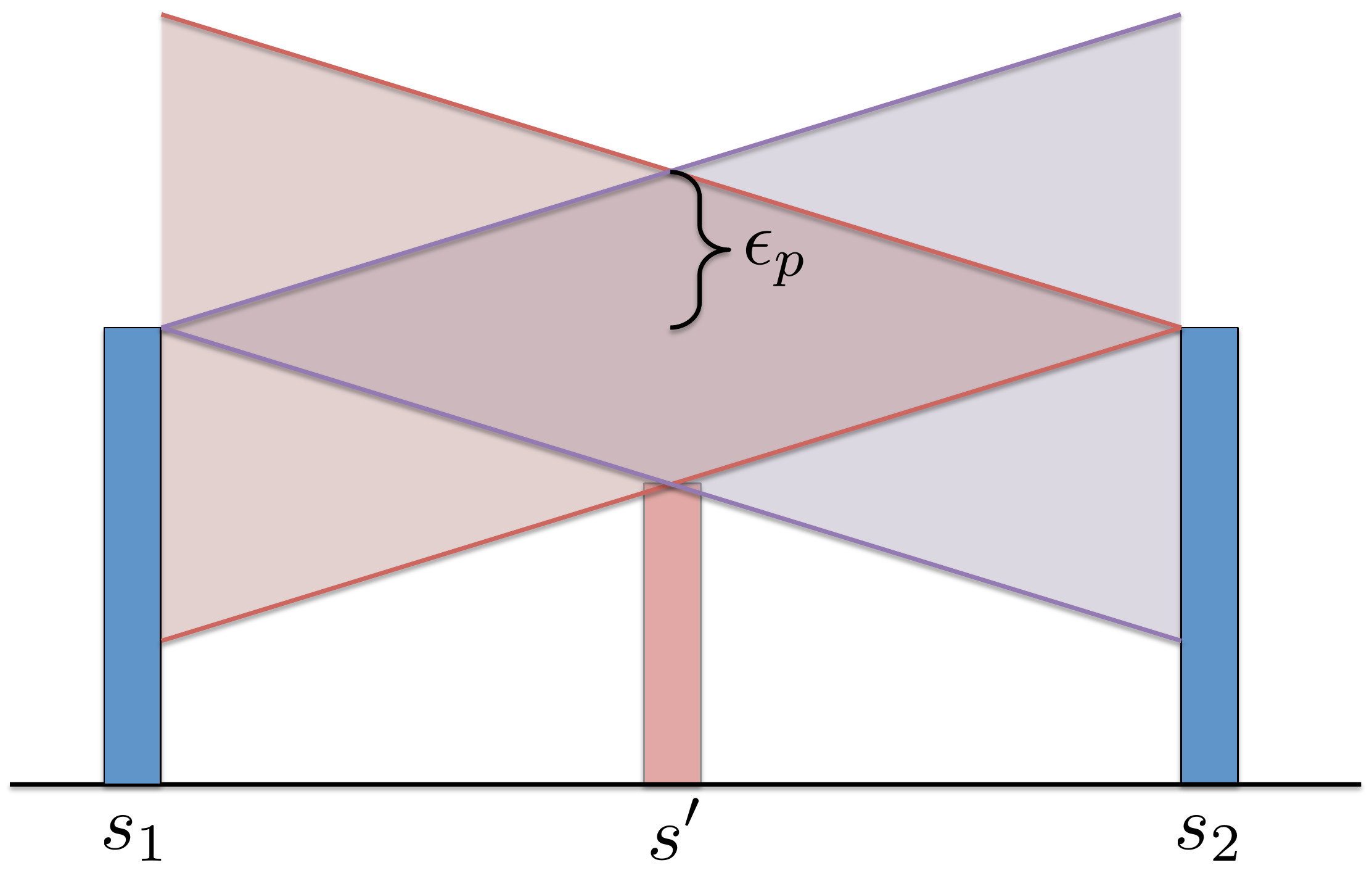}
\end{center}
\caption{An illustration of Lemma 2.  The blue bars are constraints at sampled
  points $s_1,s_2\in\sS$.  The red and purple lines indicate the maximum rate
  of change of the value function, given our settings of $\psi, \delta_\phi,
  \delta_R,$ and $\delta_P$.  The center diamond is therefore the feasible
  area for the approximate value function, and the red bar is the
  \emph{implied} constraint at some novel point $s'\in\sS$.  Because
  $\epsilon_p=\delta_\Phi\psi+\delta_R+\delta_P\psi$ is the maximum change, we
  see that the difference between the best possible setting of $\Phi(s') w$
  and the worst possible setting of
$\Phi(s') w$ is at most $2\epsilon_p$.}
\label{fig:epsilonp}
\end{figure}

\begin{lemma}
\label{lem:difVal}
Let $M_1$ and $M_2$ be MDPs that differ only in their reward vectors $R_1$ and
$R_2$.  Let $V_1^*$ and $V_2^*$ be their optimal value functions.  Then, for
$\norm{R_1-R_2}_\infty\leq\delta$,
$\norm{V_1^*-V_2^*}_\infty\leq\frac{\delta}{1-\gamma}$.
\end{lemma}

\begin{proof}Let $s$ be an arbitrary point in the sets $\sS_1$ and $\sS_2$,
  and define $r_{1i}(s)$ and $r_{2i}(s)$ to be the $i$-th reward received in
  exploring $M_1$ and $M_2$ from state $s$, respectively.  Note that
\begin{equation*}
V_1^*(s)=\sum_{i=0}^\infty\gamma^i\expect{r_{1i}(s)}
\end{equation*}
and
\begin{align*}
V_2^*(s)\leq&\sum_{i=0}^\infty\left(\gamma^i\expect{r_{1i}(s)+\delta}\right)\\
=&\sum_{i=0}^\infty\gamma^i\expect{r_{1i}(s)}+\sum_{i=0}^\infty\gamma^i\delta
\end{align*}
Therefore,
\begin{align*}
|V_1^*(s)-V_2^*(s)|\leq&\sum_{i=0}^\infty\gamma^i\expect{r_{1i}(s)}-
\left(\sum_{i=0}^\infty\gamma^i\expect{r_{1i}(s)}+
\sum_{i=0}^\infty\gamma^i\delta\right)\\
=&\sum_{i=0}^\infty\gamma^i\delta\\
=&\frac{\delta}{1-\gamma}
\end{align*}
Because this is true for an arbitrary $s$,
$\norm{V_1^*-V_2^*}_\infty\leq\frac{\delta}{1-\gamma}$.
\end{proof}

Lemma 2 is trivially proven by combining Lemmas \ref{lem:reward} and
\ref{lem:difVal}, and is illustrated by Figure \ref{fig:epsilonp}.

\end{document}